\documentclass{article}

\usepackage{arxiv}
\usepackage[utf8]{inputenc} 
\usepackage[T1]{fontenc}    
\usepackage{hyperref}       
\usepackage{url}            
\usepackage{booktabs}       
\usepackage{amsfonts}       
\usepackage{nicefrac}       
\usepackage{microtype}      
\usepackage{lipsum}		
\usepackage{graphicx}
\usepackage{doi}

\usepackage{researchpack}
\usepackage{xspace}
\usepackage[dvipsnames]{xcolor}
\usepackage[numbers,compress]{natbib}
\usepackage[capitalise,nameinlink]{cleveref}
\usepackage{enumitem}
\usepackage{tikz}
\usetikzlibrary{positioning,shapes,shadows,arrows}

\hypersetup{
    colorlinks=true,
    citecolor=Cerulean,
    linkcolor=Cerulean,
    filecolor=magenta,      
    urlcolor=magenta,
}

\newcommand{\eg}{\textit{e.g.},\xspace}

\newcommand{\etal}{et al.\@\xspace}

\title{Leveraging Explanations in Interactive Machine Learning: An Overview}

\author{
    Stefano Teso \\
	University of Trento \\
	Trento, Italy \\
	\texttt{stefano.teso@unitn.it} \\
	\And
	Öznur Alkan \\
	Optum \\
	Dublin, Ireland \\
	\texttt{oznur.alkan@optum.com} \\
	\AND
	Wolfang Stammer \\
	Technical University Darmstadt \\
	Darmstadt, Germany \\
	\texttt{wolfgang.stammer@cs.tu-darmstadt.de} \\
	\And
	Elizabeth Daly \\
	IBM Research \\
	Dublin, Ireland \\
	\texttt{elizabeth.daly@ie.ibm.com} \\
}

\renewcommand{\headeright}{}
\renewcommand{\undertitle}{}
\renewcommand{\shorttitle}{Leveraging Explanations in Interactive ML}

\hypersetup{
pdftitle={Leveraging Explanations in Interactive Machine Learning: An Overview},
pdfauthor={Stefano Teso, Öznur Alkan, Wolfgang Stammer, Elizabeth Daly},
pdfkeywords={Explainable AI, Human-in-the-loop, Interactive Machine Learning, Interactive Machine Teaching, Active Learning, Recommendation},
}

\begin{document}
\maketitle

\begin{abstract}
	Explanations have gained an increasing level of interest in the AI and Machine Learning (ML) communities in order to improve model transparency and allow users to form a mental model of a trained ML model.
	However, explanations can go beyond this one way communication as a mechanism to elicit user control, because once users understand, they can then provide feedback.
	The goal of this paper is to present an overview of research where explanations are combined with interactive capabilities as a mean to learn new models from scratch and to edit and debug existing ones.
	To this end, we draw a conceptual map of the state-of-the-art, grouping relevant approaches based on their intended purpose and on how they structure the interaction, highlighting similarities and differences between them.
	We also discuss open research issues and outline possible directions forward, with the hope of spurring further research on this blooming research topic.
\end{abstract}

\keywords{Human-in-the-Loop \and Explainable AI \and Interactive Machine Learning \and Debugging \and Model Editing}

\section{Introduction}
\label{sec:introduction}

%
%
The fields of eXplainable Artificial Intelligence (XAI) and Interactive Machine Learning (IML) have traditionally been explored separately.
%
%
On the one hand, XAI aims at making AI and Machine Learning (ML) systems more transparent and understandable, chiefly by equipping them with algorithms for explaining their own decisions~\cite{guidotti2018survey,ras2022explainable}.  Such explanations are instrumental for enabling stakeholders to inspect the system's knowledge and reasoning patterns, however stakeholders only participate as \textit{passive observers} and have no control over the system or its behavior.
%
%
On the other hand, IML focuses primarily on communication between machines and humans, and it is specifically concerned with eliciting and incorporating human feedback into the training process via intelligent user interfaces~\citep{fails2003interactive, amershi2014power, michael2020interactive, WARE2001281, HE20169, WinNT}.  IML covers a broad range of techniques for in-the-loop interaction between humans and machines, however, most research \textit{does not explicitly consider explanations}.

%
%
Recently, a number of works have sought integrating techniques from XAI within the IML loop.
The core observation behind this line of research is that, \textit{interacting through explanations} is an elegant and human-centric solution to the problem of acquiring rich human feedback, and therefore leads to higher-quality AI and ML systems, in a manner that is effective and transparent for both users and machines.
%
%
In order to accomplish this vision, these works leverage either \textit{machine explanations} obtained using techniques from XAI, \textit{human explanations} provided as feedback by sufficiently expert annotators, or both, to define and implement a suitable interaction protocol.

%
%
These two types of explanations play different roles.
By observing the machine's explanations, users have the opportunity of building a better \textit{understanding} of the machine's overall logic, which not only facilitates trust calibration~\citep{amershi2014power}, but also supports and amplifies our natural capacity of providing appropriate feedback~\citep{kulesza2015principles}.
Machine explanations are also key for identifying imperfections and bugs affecting ML models, such as reliance on confounded features that
are not causally related with the desired outcome~\citep{lapuschkin2019unmasking,geirhos2020shortcut,schramowski2020making}.
%
%
At the same time, human explanations are a very rich source of supervision for models~\citep{camburu2018snli}
and are also very natural for us to provide.  In fact, explanations tap directly into our innate learning and teaching abilities~\citep{lombrozo2006structure,mac2018teaching} and are often spontaneously provided by human annotators when given the chance~\citep{stumpf2007toward}.
%
%
Machine explanations and human feedback can also be combined to build interactive \textit{model editing} and \textit{debugging} facilities, because -- once aware of limitations or bugs in the model -- users can indicate effective improvements and supply corrective feedback~\citep{kulesza2015principles,teso2019explanatory}.

%
%
The goal of this paper is to provide a general overview and perspective of research on leveraging explanations in IML.
Our contribution is two-fold.
First, we survey a number of relevant approaches, grouping them based on their intended purpose and how they structure the interaction.
Our second contribution is a discussion of open research issues, on both the algorithmic and human sides of the problem, and possible directions forward.
This paper is not meant as an exhaustive commentary of all related work on explainability or on interactivity.
Rather, we aim to offer a conceptual guide for practitioners to understand core principles and key approaches in this flourishing research area.
In contrast to existing overviews on using explanations to guide learning algorithms~\citep{hase2021can}, debug ML models~\citep{lertvittayakumjorn2021explanation}, and transparently recommend products~\citep{zhang2020explainable}, we specifically focus on \textit{human-in-the-loop} scenarios and consider a broader range of applications, highlighting the variety of mechanisms and protocols for producing and eliciting explanations in IML.

\paragraph{Outline:}  This paper is structured as follows.
In~\cref{sec:motivation} we discuss a general recipe for integrating explanations into interactive ML, recall core principles for designing successful interaction strategies, and introduce a classification of existing approaches.
Then we discuss key approaches in more detail, organizing them based on their intended purpose and the kind of machine explanations they leverage.
Specifically, we survey methods
for debugging ML models using saliency maps and other local explanations in~\cref{sec:interacting-via-local-explanations},
for editing models using global explanations (e.g., rules) in \cref{sec:interacting-via-global-explanations}, and
for learning and debugging ML models with concept-level explanations in \cref{sec:concept-based-interaction}.
Finally, we outline remaining open problems in \cref{sec:open-problems} and related topics in \cref{sec:related-topics}.
\section{Explanations in Interactive Machine Learning}
\label{sec:motivation}

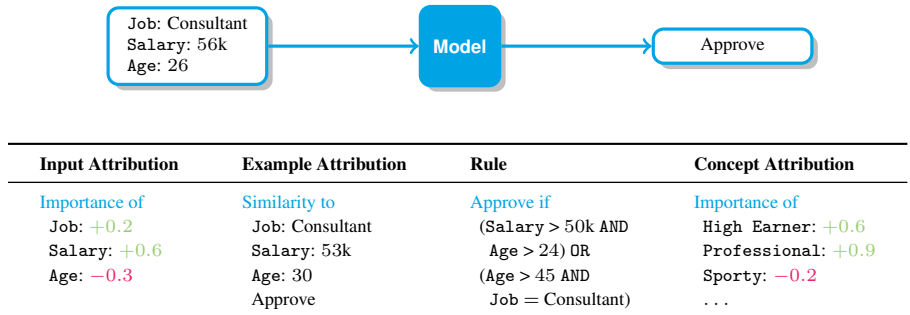
\begin{figure}[!t]

    \tikzstyle{variable-node}=[
        rectangle,
        draw=Cerulean,
        rounded corners,
        very thick,
        drop shadow,
        fill=white,
        text=black,
        text centered,
        minimum width=6em
    ]
    \tikzstyle{function-node}=[
        rectangle,
        draw=Cerulean,
        rounded corners,
        very thick,
        drop shadow,
        fill=Cerulean,
        text=white,
        text centered,
        minimum width=3em,
        minimum height=3em
    ]
    \tikzstyle{myarrow}=[
        ->,
        draw=Cerulean,
        very thick
    ]

    \begin{center}

        \begin{tikzpicture}[node distance=2cm, font=\scriptsize]
            \node (Input) [variable-node, align=left]{
                {\tt Job}: Consultant\\
                {\tt Salary}: $56$k\\
                {\tt Age}: $26$
            };
            \node (Model) [function-node, right=of Input]{
                \textbf{\textsf{Model}}
            };
            \node (Output) [variable-node, right=of Model]{
                Approve
            };
            \draw[myarrow] (Input.east) -- (Model.west);
            \draw[myarrow] (Model.east) -- (Output.west);
        \end{tikzpicture}

        \vspace{2em}
        
        \newcommand{\mytab}{\hspace{0.5em}}

        {\scriptsize
        \setlength{\tabcolsep}{12pt}
        \renewcommand{\arraystretch}{1.15}
        \begin{tabular}{llll}
            \toprule
            {\bf Input Attribution}
                & {\bf Example Attribution}
                & {\bf Rule}
                & {\bf Concept Attribution}
            \\
            \midrule
            \textcolor{Cerulean}{Importance of}
                & \textcolor{Cerulean}{Similarity to}
                & \textcolor{Cerulean}{Approve if}
                & \textcolor{Cerulean}{Importance of}
            \\
            \mytab {\tt Job}: \textcolor{YellowGreen}{$+0.2$}
                & \mytab {\tt Job}: Consultant
                & \mytab ({\tt Salary} > $50$k {\tt AND}
                & \mytab {\tt High Earner}: \textcolor{YellowGreen}{$+0.6$}
            \\
            \mytab {\tt Salary}: \textcolor{YellowGreen}{$+0.6$}
                & \mytab {\tt Salary}: $53$k
                & \mytab \; {\tt Age} > $24$) {\tt OR}
                & \mytab {\tt Professional}: \textcolor{YellowGreen}{$+0.9$}
            \\
            \mytab {\tt Age}: \textcolor{WildStrawberry}{$-0.3$}
                & \mytab {\tt Age}: $30$
                & \mytab ({\tt Age} > $45$ {\tt AND}
                & \mytab {\tt Sporty}: \textcolor{WildStrawberry}{$-0.2$}
            \\
                & \mytab Approve
                & \mytab \; {\tt Job} $=$ Consultant)
                & \mytab $\ldots$
            \\
            \bottomrule
        \end{tabular}
        }

    \end{center}

    \caption{Illustration of how various kinds of explanations support human understanding in the context of loan requests.  From left to right: input attributions and example attributions (discussed in~\cref{sec:interacting-via-local-explanations}), rules (discussed in~\cref{sec:interacting-via-global-explanations}), and concept attributions (discussed in~\cref{sec:concept-based-interaction}).}
    \label{fig:explanations}
\end{figure}

%
%
Interactive Machine Learning (IML) stands for the design and the implementation of algorithms and user interfaces in order to engage users to actually build ML models. This stands in contrast to standard procedures, in which building a model is a fully automated process and domain experts have little control beyond data preparation~\citep{WARE2001281}.
%
%
Research in IML explores ways to learn and manipulate models through an intuitive human-computer interface~\citep{michael2020interactive} and encompasses a variety of learning and interaction strategies.
%
%
Perhaps the most well-known IML framework is active learning~\citep{settles2012active,herde2021survey}, which tackles learning high-performance predictors in settings in which supervision is expensive.  To this end, active learning algorithms interleave acquiring labels of carefully selected unlabeled instances from an annotator and model updates.
%
%
As another brief example, consider a recommender solution in a domain like movies, videos, or music, where the user can provide explicit feedback through rating the recommended items~\citep{HE20169}. These ratings can then be infused to the recommendation model's decision making process so as to tailor the recommendations towards end users interests.

%
%
In IML, users are encouraged to 
shape the decision making process of the model, so it is important for users to build a correct mental model of the “intelligent agent”, which will then allow them to seamlessly interact with it.
%
%
Conversely, since user feedback and involvement are so central, uninformed feedback risks wasting annotation effort and ultimately compromising model quality.
%
%
Not all approaches to IML are equally concerned with user understanding.
For instance, in active learning the machine is essentially a black-box, with no information being disclosed about what knowledge it has acquired and what effect feedback has on it~\citep{teso2019explanatory}.
Strategies for interactive customization of ML models, like the one proposed by \citet{fails2003interactive}, are less opaque, in that users can explore the impact of their changes and tune their feedback accordingly.  Yet, the model's logic can only be (approximately) reconstructed from changes in behavior, making it hard to \textit{anticipate} what information should be provided to guide the model in a desirable direction~\citep{kulesza2015principles}.
%
%
Following \citet{kulesza2015principles}, we argue that proper interaction requires transparency and an understanding of the underlying model's logic.  And it is exactly here that \textit{explanations} can be used to facilitate this process.

\subsection{A General Approach for Leveraging Explanations in Interaction}

%
%
In order to appreciate the potential roles played by explanations, it is instructive to look at \textit{explanatory debugging}~\citep{kulesza2010explanatory,kulesza2015principles}, the first framework to explicitly leverage them in IML, and specifically at EluciDebug~\citep{kulesza2015principles}, its proof-of-concept implementation. 

EluciDebug is tailored for interactive customization of Na\"ive Bayes classifiers in the context of email categorization.
To this end, it presents users with explanations that illustrate the relative contributions of the model's prior and likelihood towards the class probabilities.  In particular, its explanations convey what \textit{words} the model uses to distinguish work emails from personal emails and how much they impact its decisions.
In a second step, the user has the option of increasing or decreasing the relevance of certain input variables toward the choice of certain classes by directly adjusting the model weights.  Continuing with our email example, the user is free to specify relevant words that the system is currently ignoring and irrelevant words that the system is wrongly relying on.
This very precise form of feedback contrasts with traditional label-based strategies, in which the user might, e.g., tell the system that a message from her colleague about baseball is a personal (rather than work-related) communication, but has no direct control over what inputs the system decides to use.
The responsibility of choosing what examples (e.g., wrongly classified emails) to provide feedback on is left to the user, and the interaction continues until she is happy with the system's behavior.
EluciDebug was shown to help users to better understand the system's logic and to more quickly tailor it toward their needs~\citep{kulesza2015principles}.

EluciDebug highlights how explanations contribute to both \textit{understanding} and \textit{control}, two key elements that will reoccur in all approaches we survey.  We briefly unpack them in the following.

\paragraph{Understanding.}  By observing the machine's explanations, users get the opportunity of building a better \textit{understanding} of the machine's overall logic.
This is instrumental in uncovering limitations and flaws in the model~\citep{VILONE202189}.
As a brief example, consider a recommender solution in a domain like movies, videos, or music, where the users are presented with explanations in the form of a list of features that are found to be most relevant to the users' previous choices~\cite{Tintarev2007}. Upon observing this information, users can see the assumptions the underlying recommender has made for their interests and preferences. It might be the case that the model made incorrect assumptions for the users' preferences possibly due to some  changes of interests which is not explicitly available in the data. In such a scenario, explanations provide a perfect ground for understanding the underlying model's behavior.

Explanations are also instrumental for identifying models that rely on confounds in the training data, such as watermarks in images, that happen to correlate with the desired outcome but that are not causal for it~\citep{lapuschkin2019unmasking,geirhos2020shortcut}.  Despite achieving high accuracy during training, these models generalize poorly to real-world data where the confound is absent.  Such buggy behavior can affect high-stakes applications like COVID-19 diagnosis~\citep{degrave2021ai} and scientific analysis~\citep{schramowski2020making}, and cannot be easily identified using standard evaluation procedures without explanations.
Ideally, the users would develop a structural mental model that gives them a deep understanding of how the model operates, however a functional understanding is often enough for them to be able to interact~\citep{sokol2020one}.
The ability of disclosing issues with the model, in turn, facilitates trust calibration~\citep{amershi2014power}.  This is especially true in interactive settings as here the user can witness how the model evolves over time, another important factor that contributes to trust~\citep{waytz2014mind,wang2016trust}.

\paragraph{Control.}  Understanding supports and amplifies our natural capacity of providing appropriate feedback~\citep{kulesza2015principles}:  once bugs and limitations are identified, interaction with the model enables end-users to modify the algorithm in order to correct those flaws.
Bi-directional communication between users and machines together with transparency enables \textit{directability}, that is, the ability to rapidly assert control or influence when things go astray~\citep{hoffman2013trust}.
Clearly, control is fundamental if one plans to take actions based on a     model's prediction, or to deploy a new model in the first place~\citep{ribeiro2016should}.
At the same time, directability also contributes to trust allocation~\citep{hoffman2013trust}.
The increased level of control can also help to achieve significant gains in the end user's satisfaction.

Human feedback can come in many forms, and one of these forms is explanations, either from scratch or by using the machine's explanations as a starting point~\citep{teso2019explanatory}.
This type of supervision is very informative:  a handful of explanations are oftentimes worth many labels, substantially reducing the sample complexity of \textit{learning} (well-behaved) models~\citep{camburu2018snli}.
Importantly, it is also very natural for human to provide as explanations lie at the heart of human communication and tap directly into our innate learning and teaching abilities~\citep{lombrozo2006structure,mac2018teaching}.  In fact, \citet{stumpf2007toward} showed that when given the chance to provide free-form annotations, users had no difficulty providing generous amounts of feedback.

\paragraph{Principles.}  To ground the exchange of explanations between an end user and a model, \citet{kulesza2010explanatory} presented a set of key principles around \textit{explainability} and \textit{correctability}. Although the principles are discussed in the context of explanatory debugging, they apply to all the approaches that are presented in this paper.  These include:
($1$) Being iterative, so as to enable end-users to build a reasonable and informed model of the machine's behavior.
($2$) Presenting sound, faithful explanations that do not over-simplify the model's reasoning process.
($3$) Providing as complete a picture of the model as possible, without omitting elements that play an important role in its decision process.
($4$) Avoiding to overwhelming the user, as this complicates understanding and feedback construction.
($5$) Ensuring that explanations are actionable, making them engaging for users to attend to, thus encouraging understanding while enabling users to adjust them to their expertise.
($6$) Making user changes easily reversible.
($7$) Always honoring feedback, because when feedback is disregarded users may stop bothering to interact with the system.
($8$) Making sure to effectively communicate what effects feedback has on the model.
Clearly there is a tension behind these principles, but they nonetheless are useful in guiding the design of explanation-based interaction protocols.  As we will discuss in \cref{sec:open-problems}, although existing approaches attempt to satisfy one or more of these desiderata, no \textit{general} method yet exists that satisfies all of them.

\begin{table}[!t]
    \centering
    \scriptsize
    \begin{tabular}{lllll}
    \toprule
    \textbf{Goal}
        & \textbf{Explanations}
        & \textbf{Feedback}
        & \textbf{Incorporation}
        & \textbf{Method}
    \\

    %
    %

    \midrule
    Learning
        & Local, CA
        & Adjust Feature Association
        & Update model w/ auxiliary loss 
        & \citet{lage2020learning}
    \\
    \cmidrule[\lightrulewidth]{3-5}
        & 
        & Adjust Encodings
        & Update model w/ auxiliary loss 
        & \citet{stammer2022interactive}
    \\

    %
    %

    \midrule
    Debugging
        & Local, IA
        & Adjust Parameters
        & Update model w/ improved parameters 
        & EluciDebug~\citep{kulesza2015principles}
    \\
    \cmidrule[\lightrulewidth]{3-5}
        &
        & Adjust Attributions
        & Update data 
        & CAIPI~\citep{teso2019explanatory}
    \\
    \cmidrule[\lightrulewidth]{4-5}
        &
        &
        & Update model w/ auxiliary loss 
        & RRR~\citep{schramowski2020making,ross2017right}
    \\
    \cmidrule[\lightrulewidth]{4-5}
        &
        &
        & Update model w/ auxiliary loss 
        & \citet{teso2019toward}
    \\
    \cmidrule[\lightrulewidth]{2-5}
        & Local, EA
        & Additional Features
        & Update model w/ additional classifiers
        & ALICE~\citep{liang2020alice}
    \\
    \cmidrule[\lightrulewidth]{3-5}
        &
        & Adjust Attributes
        & Update data 
        & \citet{Biswas2013}
    \\
    \cmidrule[\lightrulewidth]{3-5}
        &
        & Example Similarity
        & Update data 
        & HILDIF~\citep{zylberajch2021hildif}
    \\
    \cmidrule[\lightrulewidth]{3-5}
        &
        & Counter Examples
        & Update data 
        & CINCER~\citep{teso2021interactive}
    \\
    \cmidrule[\lightrulewidth]{2-5}
        & Local, CA
        & Adjust Attributions
        & Update model w/ auxiliary loss 
        & RRC~\citep{stammer2021right}
    \\
    \cmidrule[\lightrulewidth]{4-5}
        &
        &
        & Update model w/ auxiliary loss 
        & \citet{bontempelli2021toward}
    \\
    \cmidrule[\lightrulewidth]{4-5}
        &
        &
        & Update model w/ auxiliary loss 
        & ProtoPDebug~\citep{bontempelli2022concept}
    \\
    \cmidrule[\lightrulewidth]{3-5}
        &
        & Sample Pairing
        & Update model w/ auxiliary loss 
        & Shao et al.~\citep{shao2022right}
    \\
    \cmidrule[\lightrulewidth]{2-5}
        & Global, Rules
        & Adjust Attributions
        & Update model w/ hard constraint 
        & FIND~\citep{lertvittayakumjorn2020find}
    \\
    \cmidrule[\lightrulewidth]{4-5}
        &
        &
        & Update model w/ auxiliary loss 
        & REMOTE~\citep{yao2021refining}
    \\
    \cmidrule[\lightrulewidth]{3-5}
        &
        & Counter Examples
        & Update data 
        & XGL~\citep{popordanoska2020machine}
    \\

    %
    %

    \midrule
     Editing
        & Global, Rules
        & Rule Editing
        & Update data 
        & FROTE~\citep{alkan2022frote}
    \\
    \cmidrule[\lightrulewidth]{4-5}
        &
        &
        & Post-processing 
        & Overlay~\citep{daly2021user}
    \\
    \cmidrule[\lightrulewidth]{4-5}
        & 
        &
        & Post-processing 
        & XIML~\citep{Lijie2022}
    \\
    \cmidrule[\lightrulewidth]{3-5}
        &
        & Adjust Feature Association
        & Update model w/ auxiliary loss 
        & \citet{Antognini2021InteractingWE}
    \\
    \cmidrule[\lightrulewidth]{4-5}
        &
        &
        & Post-processing 
        & \citet{Alkan2019, IRF_CSCW_2021}
    \\
    \bottomrule
\end{tabular}
    \vspace{1.5em}
    \caption{Table of methods covered in this overview.  We here differentiate the various methods in their algorithmic goals (Goal), the type of explanations they consume (Explanations), the type of feedback provided by the user (Feedback), and, finally, the strategy of incorporating the explanatory user feedback (Incorporation). \textit{Abbreviations}: CA $=$ concept attribution, EA $=$ example attribution, IA $=$ input attribution.}
    \label{tab:methods}
\end{table}

\subsection{Dimensions of Explanations in Interactive Machine Learning}
\label{sec:dimensions}

Identifying \textit{interaction} and \textit{explainability} as two key capabilities of a well performing \textit{and} trust-worthy ML system, motivates us to layout this overview on leveraging explanations in interactive ML.
The methods we survey tackle different applications using a wide variety of strategies.  In order to identify common themes and highlight differences, we organize them along four dimensions:

\noindent
\textbf{Algorithmic goal}:  We identify three high-level scenarios.
One is that of using explanation-based feedback, optionally accompanied by other forms of supervision, to \textit{learn} an ML model from scratch.  Here, the machine is typically in charge of asking appropriate questions, feedback may be imperfect, and the model is updated incrementally as feedback is received.
Another scenario is model \textit{editing}, in which domain experts are in charge of inspecting the internals of a (partially) trained model (either directly if the model is white-box or indirectly through its explanations) and can manipulate them to improve and expand the model. Here feedback is typically assumed high-quality and used to constrain the model's behavior.
The last scenario is \textit{debugging}, where the focus is on fixing issues of misbehaving (typically black-box) models and the machine's explanations are used to both spot bugs and elicit corrective feedback.
Naturally, there is some overlap between goals.  Still, we opt to keep them separate as they are often tackled using different algorithmic and interaction strategies.

\noindent
\textbf{Type of machine explanations}:  The approaches we survey integrate four kinds of machine explanations:  \textit{input attributions} (IAs), \textit{example attributions} (EAs), \textit{rules}, and \textit{concept attributions} (CAs).
IAs identify those input variables that are responsible for a given decision, and as such they are \textit{local} in nature.  EAs and CAs are also local, but justify decisions in terms of relevant training examples and high-level concepts, respectively.  At the other end of the spectrum, rules are \textit{global} explanations in that they aim to summarize, in an interpretable manner, the logic of a whole model.
These four types of explanations are illustrated in~\cref{fig:explanations} and described in more detail in the next sections

\noindent
\textbf{Type of human feedback and incorporation strategy}:  Algorithmic goal and choice of machine explanations act as a determiner for the types of interactions that can happen, in turn affecting two other important dimensions, namely the \textit{type of feedback} that can be collected and the way the machine can \textit{consume this feedback}~\citep{narayanan2018humans}.
Feedback ranges from updated parameter values, as in EluciDebug, to additional data points, to gold standard explanations supplied by domain experts.
Incorporation strategies go hand-in-hand, and range from updating the model's parameters as instructed to (incrementally) retraining the model, perhaps including additional loss terms to incorporate explanatory feedback.
All details are given in the following sections.

%
%
The methods we survey are listed in~\cref{tab:methods}.
Notice that the two most critical dimensions, namely algorithmic goal and type of machine explanations, are tightly correlated:  learning approaches tend to rely on local explanations and editing approaches on rules, while debugging approaches employ both.
For this reason, we chose to structure the next three sections by explanation type.
One final remark before proceeding.  Some of the approaches we cover rely on choosing specific instances or examples to be presented to the annotator.  Among them, some rely on \textit{machine-initiated} interaction, in the sense that they leave this choice to the machine (for instance, methods grounded on active learning tend to pick specific instances that the model is uncertain about~\citep{settles2012active}), while others rely on \textit{human-initiated} interaction and expect the user to pick instances of interest from a (larger) set of options.  We do not group approaches based on this distinction, so as to keep our categorization manageable.  The specific type of interaction used will be made clear in the following sections on a per-method basis.
\section{Interacting via Local Explanations}
\label{sec:interacting-via-local-explanations}

In this section, we discuss IML approaches that rely on \textit{local explanations} to carry out interactive model debugging.
Despite sharing some aspects with EluciDebug~\citep{kulesza2015principles}, these approaches exploit modern XAI techniques to support state-of-the-art ML models and explore alternative interaction protocols.
Before reviewing them, we briefly summarize those types of local explanations that they build on.

\subsection{Input Attributions and Example Attributions}

Given a classifier and a target decision, local explanations identify a subset of ``explanatory variables'' that are most responsible for the observed outcome.  Different types of local explanations differ in what variables they consider and in how they define responsibility.

\begin{figure}[!t]
    \centering
    \begin{tabular}{c|c}
        \includegraphics[width=0.45\textwidth]{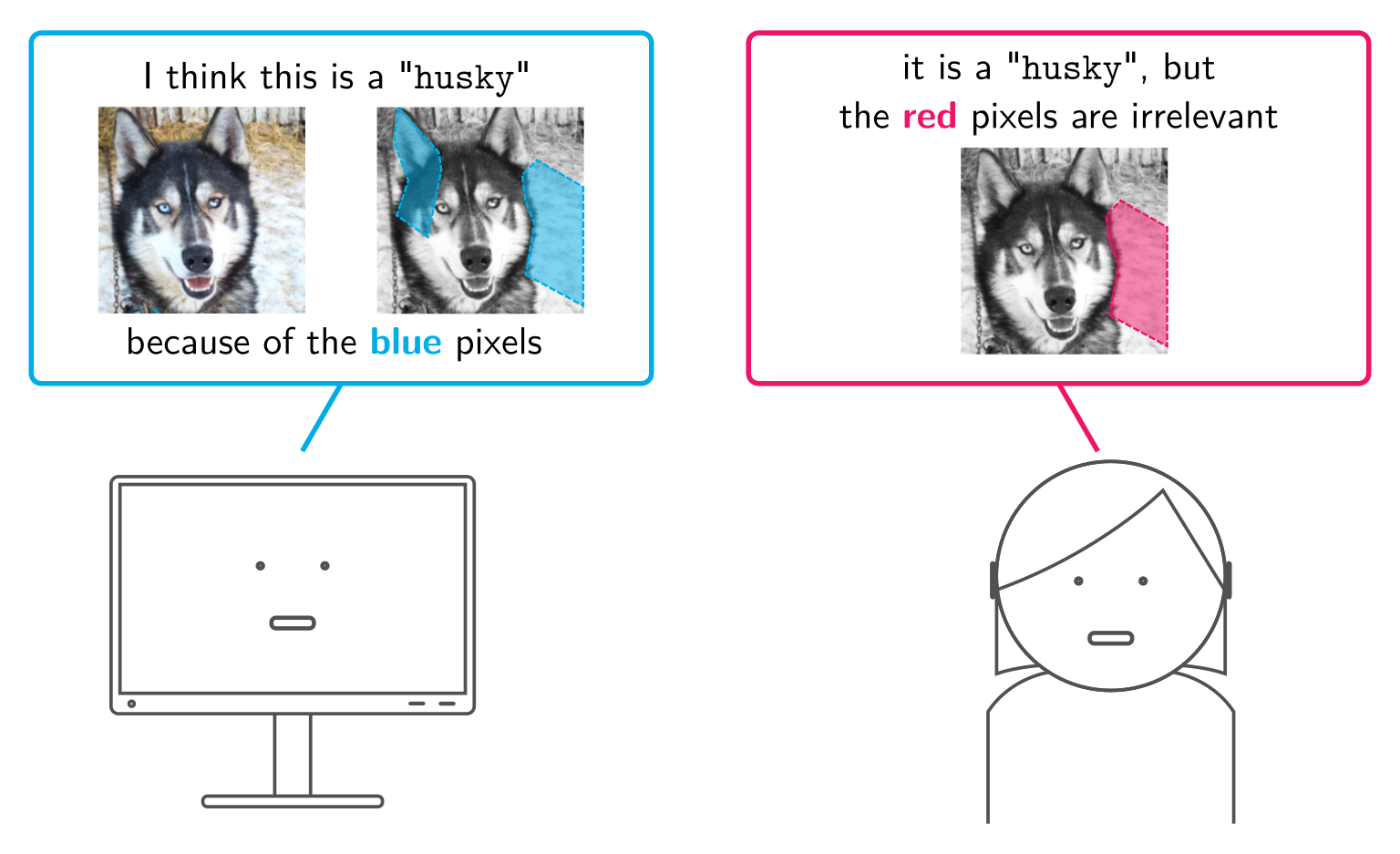}
        & \includegraphics[width=0.45\textwidth]{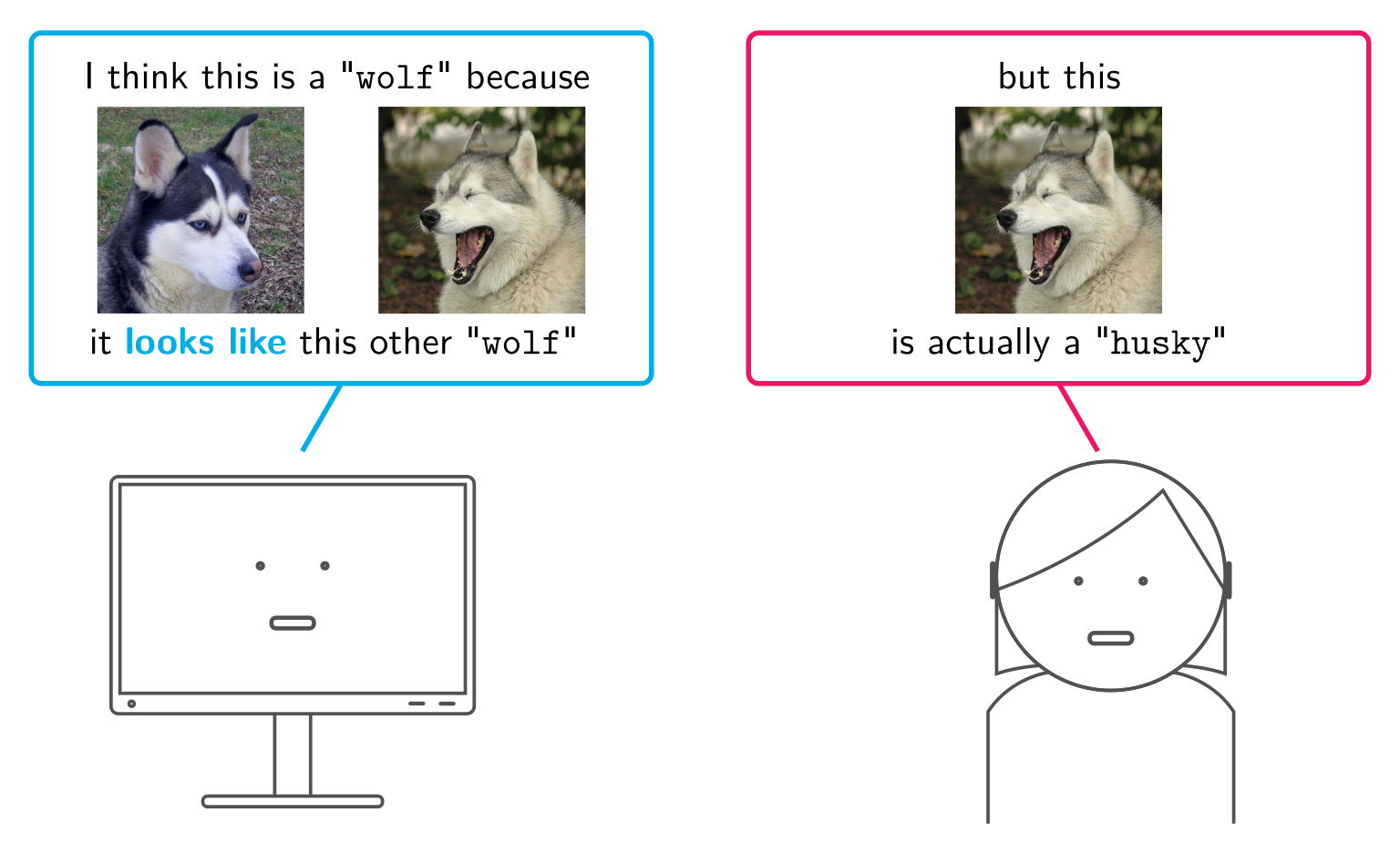}
    \end{tabular}
    \caption{Illustration of two explanation strategies based on local explanations.
    \textbf{Left}: The machine explains its predictions by highlighting relevant \textit{input variables} -- in this case, relevant pixels -- and the user replies with an improved attribution map.
    \textbf{Right}: The machine justifies its predictions in terms of training examples that support them, and the user either corrects the associated label.
    In both cases, the data and model are aligned to the user's feedback.}
    \label{fig:interacting-with-local-explanations}
\end{figure}

\textit{Input attributions}, also known as saliency maps, convey information about relevant vs.\@ irrelevant input variables.
For instance, in loan request approval an input attribution might report the relative importance of variables like {\tt Job}, {\tt Salary} and {\tt Age} of the applicant, as illustrated in~\cref{fig:explanations}, and in image tagging that of subsets of pixels, as shown in~\cref{fig:interacting-with-local-explanations} (left).
A variety of attribution algorithms have been developed.
Gradient-based approaches like Input Gradients (IGs)~\citep{baehrens2010explain,simonyan14deep}, GradCAM~\citep{selvaraju2017grad}, and Integrated Gradients~\citep{sundararajan2017axiomatic} construct a saliency map by measuring how sensitive the score or probability assigned by the classifer to the decision is to perturbations of individual input variables.
This information is read off from the gradient of the model's output with respect to its input, and as such it is only meaningful if the model is differentiable and the inputs are continuous (e.g., images).
Sampling-based approaches like LIME~\citep{ribeiro2016should} and SHAP~\citep{vstrumbelj2014explaining,lundberg2017unified} capture similar information~\citep{garreau2020explaining}, but they rely on sampling techniques that are applicable also to non-differentiable models and to categorical and tabular data.
For instance, LIME distills an interpretable surrogate model using random samples labeled by the black-box predictor and then extracts input relevance information from the surrogate.
Despite their wide applicability, these approaches tend to be more computationally expensive than gradient-based alternatives~\citep{van2021tractability} and, due to variance inherent in the sampling step, in some cases their explanations may portray an imprecise picture of the model's decision making process~\citep{zhang2019should,teso2019toward}.
One last group of approaches focus on identifying the smallest subset of input variables whose value, once fixed, ensures that the model's output remains the same regardless of the value taken by the remaining variables~\citep{shih2018symbolic,wang2020towards,camburu2020struggles}, and typically come with faithfulness guarantees.  Although principled, these approaches however have not yet been used in explanatory interaction.

\textit{Example attributions}, on the other hand, explain a target decisions in terms of those training examples that most contributed to it, and are especially natural in settings, like medical diagnosis, in which domain experts are trained to perform case-based reasoning~\citep{bien2011prototype,kim2014bayesian,chen2019looks}.
For instance, in~\cref{fig:explanations} a loan application is approved by the machine because it is similar to an previously approved application and in~\cref{fig:interacting-with-local-explanations} (right) a mislabeled training image fools the model into mispredicting a {\tt husky} dog as a {\tt wolf}.
For same classes of models, like nearest neighbor classifiers and prototype-based predictors, example relevance can be easily obtained.
For all other models, it can in principle be evaluated by removing the example under examination from the training set and checking how this changes the models' prediction upon retraining.  This na\"ive solution however scales poorly, especially for larger models for which retraining is time consuming.
A more convenient alternative are Influence Functions (IFs), which offer an efficient strategy for approximating example relevance without retraining~\citep{koh2017understanding}, yielding a substantial speed-up.
Evaluating IFs is however non-trivial, as it involves inverting the Hessian of the model, and can be sensitive to factors such as model complexity~\citep{basu2020influence} and noise~\citep{teso2021interactive}.
This has prompted researchers to develop more scalable and robust algorithms for IFs~\citep{guo2020fastif} as well as alternative strategies to approximate example relevance~\citep{yeh2018representer,khanna2019interpreting}.

\subsection{Interacting via Input Attributions}

%
%
One line of work on integrating input attributions and interaction is \textit{eXplanatory Interactive Learning} (XIL)~\citep{teso2019explanatory,schramowski2020making}.
In the simplest case, XIL follows the standard active learning loop~\citep{settles2012active}, except that whenever the machine queries the label of a query instance, it also presents a \textit{prediction} for that instance and a \textit{local explanation} for the prediction.
Contrary to EluciDebug, which is designed for interpretable classifiers, in XIL the model is usually a black-box, e.g., a support vector machine or a deep neural network, and explanations are extracted using input attribution methods like LIME~\citep{teso2019explanatory} or GradCAM~\citep{schramowski2020making}.
At this point, the annotator supplies a label for the query instance -- as in regular active learning -- and, optionally, corrective feedback on the explanation.   The user can, for instance, indicate what input variables the machine is wrongly relying on for making its prediction. 
Consider~\cref{fig:interacting-with-local-explanations} (left):  here the query instance depicts a {\tt husky} dog, but the model wrongly classifies it as a {\tt wolf} based on the presence of snow in the background (in \textbf{\textcolor{Cerulean}{blue}}), which happens to correlate with wolf images in the training data.  To correct for this, the user indicates that the snow should not be used for prediction (in \textbf{\textcolor{WildStrawberry}{red}}).

XIL then aligns the model based on this corrective feedback.
CAIPI~\citep{teso2019explanatory}, the original implementation of XIL, achieves this using data augmentation.  Specifically, CAIPI makes a few copies of the target instance (e.g., the husky image in \cref{fig:interacting-with-local-explanations}) and then \textit{randomizes} the input variables indicated as irrelevant by the user while leaving the label unchanged, yielding a small set of synthetic examples.  These are then added to the training set and the model is retrained.  Essentially, this teaches the model to classify the image correctly \textit{without relying on the randomized variables}.
Data augmentation proved effective at debugging shallow models for text classification and other tasks~\citep{teso2019explanatory,slany2022caipi} and at reducing labeling effort~\citep{slany2022caipi}, at the cost of requiring extra space to store the synthetic examples.

A more refined version of CAIPI~\citep{schramowski2020making} solves this issue by introducing two improvements:  LIME is replaced with GradCAM~\citep{selvaraju2017grad}, thus avoiding sampling altogether, and the model is aligned using a generalization of the \textit{right for the right reasons} (RRR)~\citep{ross2017right} modified to work with GradCAM.
Essentially, the RRR loss penalizes the model proportionally to the relevance that its explanations assign to inputs that have been indicated as irrelevant by the user.
Combining it with a regular loss for classification (e.g., the categorical cross-entropy loss) yields an end-to-end differentiable training pipeline that can be optimized using regular back-propagation.  This in turn leads to shorter training times, especially for larger models, without the need for extra storage space.
This approach was empirically shown to successfully avoid Clever Hans behavior in deep neural networks used for hyperspectral analysis of plant phenotyping data~\citep{schramowski2020making}.

Several alternatives to the RRR loss have been developed.
The Contextual Decomposition Explanation Penalization (CDEP)~\citep{rieger2020interpretations} follows the same recipe, but it builds on Contextual Decomposition~\citep{singh2018hierarchical}, an attribution technique that takes relationships between input variables into account.
The approach of~\citet{yao2021refining} enables annotators to dynamically explore the space of feature interactions and fine-tune the model accordingly.
The Right for Better Reasons (RBR) loss~\citep{shao2021right} improves on gradient-based attributions by replacing input gradients with their influence function~\citep{koh2017understanding}, and it was shown to be more robust to changes in the model and speed up convergence.
Human Importance-aware Network Tuning (HINT)~\citep{selvaraju2019taking} takes a different route in that it rewards the model for activating on inputs deemed \textit{relevant} by the annotator.
Finally, \citet{teso2019toward} introduced a ranking loss designed to work with partial and possibly sub-optimal explanation corrections.
These methods have recently been compared in the context of XIL by~\citet{friedrich2022typology}. There, the authors introduce a set of benchmarking metrics and tasks and conclude that the ``no free lunch'' theorem \cite{wolpert1997no} also holds for XIL, i.e., no method exceeds on all evaluations.

ALICE~\citep{liang2020alice} also augments active learning, but it relies on contrastive explanations.  In each interaction round, the machine selects a handful of class \textit{pairs}, focusing on classes that the model cannot discriminate well, and for each of them asks an annotator to provide a natural language description of what features enable them to distinguish between the two classes.  It then uses semantic parsing to convert the feedback into rules and integrates it by ``morphing'' the model architecture accordingly.
\citet{parkash2012attributes,Biswas2013}, on the other hand, enable users to specify what attributes make an instance a negative, and use them to acquire negative examples using pre-trained attribute classifiers.

FIND is an alternative approach for interactively debugging models for natural language processing tasks~\citep{lertvittayakumjorn2020find}.
What sets it apart is that interaction is framed in terms of \textit{sets} of local explanations.
FIND builds on the observation that, by construction, local explanations fail to capture how the model behaves in regions far away from the instances for which the user receives explanations.  This, in turn, complicates acquiring high-quality supervision and allocating trust~\citep{wu2019local,popordanoska2020machine}.
FIND addresses this issue by extracting those words and $n$-grams that best characterize each latent feature acquired by the model, and then visualizing the relationship between words and features using a word cloud.  The characteristic words are obtained using layer-wise relevance propagation~\citep{bach2015pixel}, a technique akin to input gradients, to all examples in the training set.
Based on this information, the user can turn off those latent features that are not relevant for the predictive task.
For instance, if a model has learned a latent feature that strongly correlates with a polar word like ``{\tt love}'' and uses it to categorize documents into non-polar classes such as spam and non-spam, FIND allows to instruct the model to no longer rely on this feature.
This is achieved by introducing a hard constraint directly into the prediction process.
Other strategies for overcoming the limits of explanation locality are discussed in~\cref{sec:interacting-via-global-explanations}.

\subsection{Interacting via Example Attributions}

Example attributions also have a role to play in interactive debugging.  Existing strategies aim to uncover and correct cases where a model relies on ``bad'' training examples for its predictions, but target different types bugs and elicit different types of feedback.
HILDIF~\citep{zylberajch2021hildif} uses a fast approximation of influence functions~\citep{guo2020fastif} to identify examples in support of a target prediction and and then asks a human-in-the-loop to verify whether the their \textit{level of influence} is justified.
It then calibrates the influence of these examples on the model via data augmentation.
Specifically, HILDIF tackles NLP tasks and it augments those training examples that are most relevant, as determined by the user, by replacing words by synonyms, effectively boosting their relative influence compared to the others.
In this sense, the general idea is reminiscent of XIL, although viewed from the lens of example influence rather than attribute relevance.

A different kind of bug occurs when a model relies on mislabeled examples, which are frequently encounteded in applications, especially when interacting with (even expert) human annotators~\citep{frenay2014classification}.
CINCER~\citep{teso2021interactive} offers a direct way to deal with this issue in sequential learning tasks.
To this end, in CINCER the machine monitors incoming examples and asks a user to double-check and re-label those examples that look suspicious~\citep{zeni2019fixing,bontempelli2020learning}.  A major issue is that the machine's \textit{skepticism} may be wrong due to -- again -- presence of mislabeled training examples.  This begs the question: is the machine skeptical for the right reasons?
CINCER solves this issue by identifying those training examples that most support the model's skepticism using IFs, giving the option of \textit{relabeling} the suspicious incoming example, the influential examples, or both.
This process is complementary to HILDIF, as it enables stakeholders to gradually improve the quality of the training data itself, and -- as a beneficial side-effect -- also to calibrate its influence on the model.

\subsection{Benefits and Limitations}

Some of the model alignment strategies employed by the approaches overviewed in this section were born for \textit{offline} alignment task~\citep{ross2017right,ghaeini2019saliency,selvaraju2019taking,hase2021can}.
A major issue of this setting is that it is not clear where the ground-truth supervision comes from, as most existing data sets do not come with dense (e.g., pixel- or word-level) relevance annotations.
In interactive the settings we consider, this information is naturally provided by a human annotator.
Another important advantage is that in interactive settings users are only required to supply feedback tailored for those buggy behaviors exhibited by the model, dramatically reducing the annotation burden.
The benefits and potential drawbacks of explanatory interaction are studied in detail by~\citet{ghai2021explainable}.
\section{Interacting via Global Explanations}
\label{sec:interacting-via-global-explanations}

All explanations discussed so far are \emph{local}, in the sense that they explain a given decision $f(\vx) = y$ of a specific input $\vx$. Local explanations enable the user to build a mental model of individual predictions, however bringing together the information gleaned by examining a number of individual predictions may prove challenging to get an overall understanding of a model.  Lifting this restriction immediately gives \emph{regional} explanations, which are guaranteed to be valid in a larger, well-defined region around $\vx$. For instance, the explanations output by decision trees are regional in this sense.
Global explanations aim to describe the behavior of $f$ across all of its domain~\citep{guidotti2018survey} providing an approximate overview of the model \citep{craven1995extracting, deng2019interpreting}. One approach is to train a directly interpretable model such as a decision tree or a rule set, using the same training data optimised for the interpretable model to behave similarly to the original model. This provides a surrogate white-box model of $f$ which can then be used as an approximate global map of $f$~\citep{guidotti2018survey, lundberg2020local}. Other approaches start with local or regional  explanations and merge them to provide a global explainer \citep{lundberg2020local, setzu2021glocalx}.

\begin{figure}[!t]
    \centering
    \begin{tabular}{c|c}
        \includegraphics[width=0.45\textwidth]{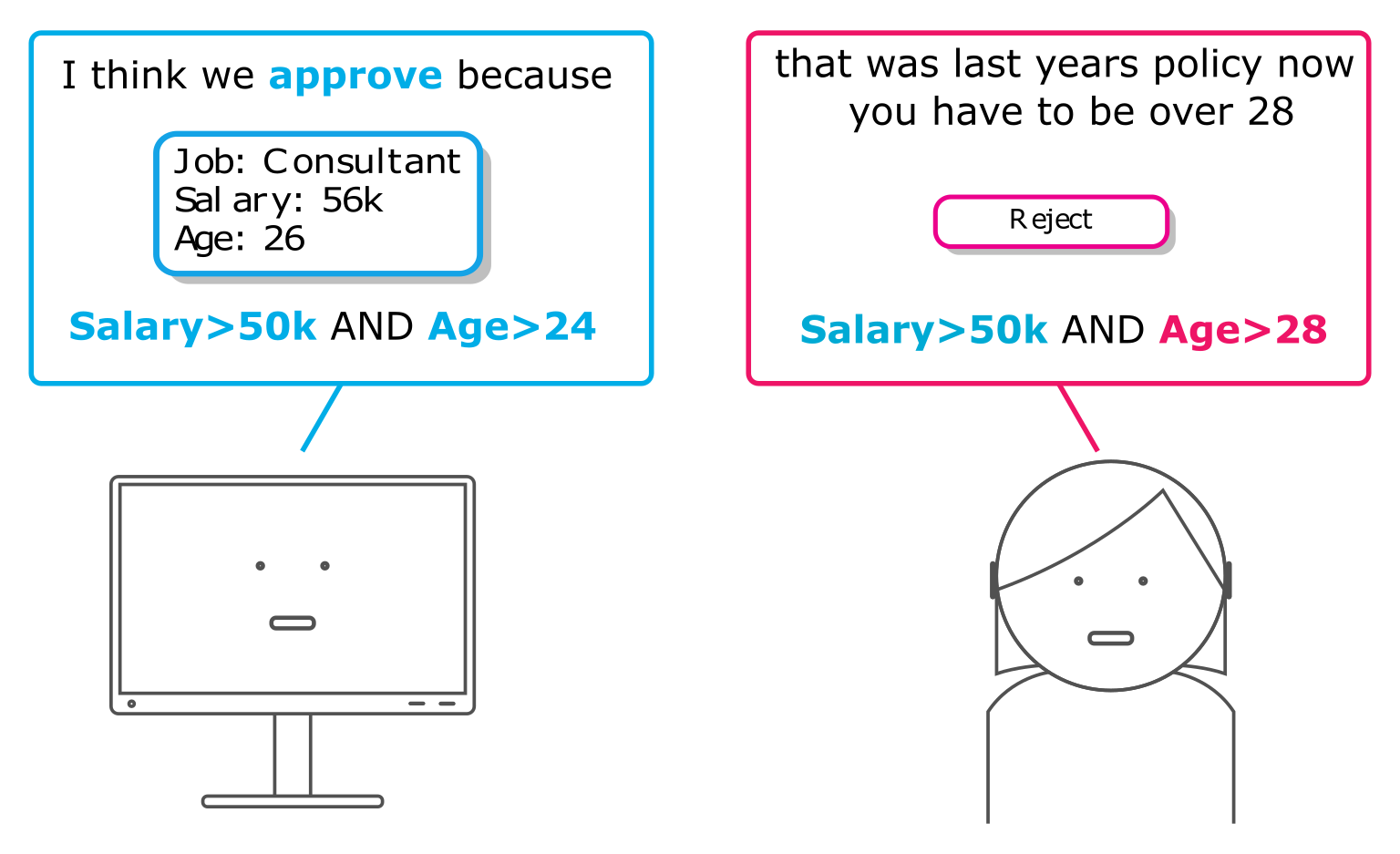}
        & \includegraphics[width=0.45\textwidth]{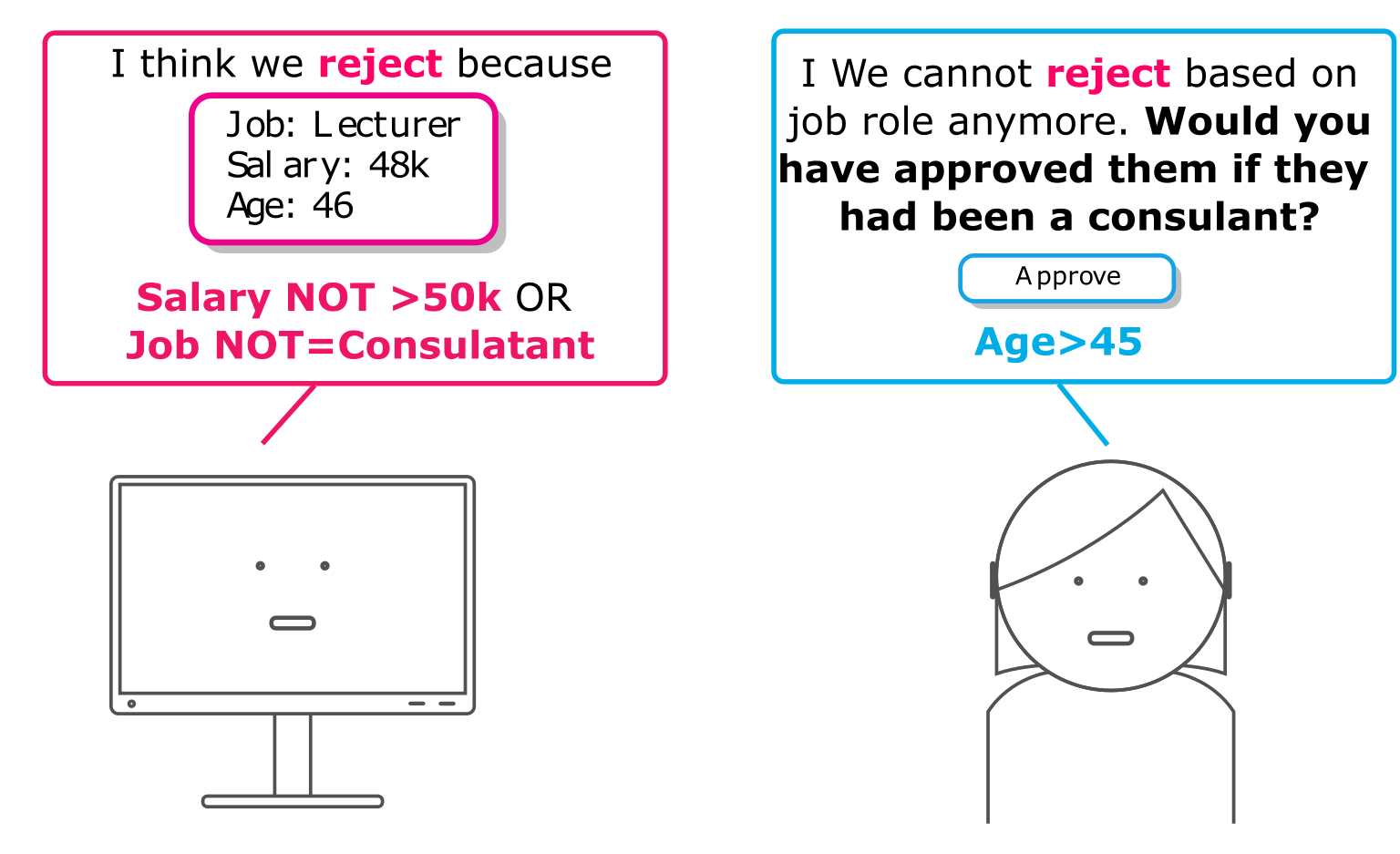}
    \end{tabular}
    \caption{Illustration of rule-based explanation feedback.
    \textbf{Left}: Rule-based explanations shown to the user to support the prediction. 
    \textbf{Right}: The input instance does not satisfy any rules to support approve, so the clauses the input instance violates can be shown so the user can validate whether those reasons should be upheld. }
    \label{fig:interacting-with-rule-explanations}
\end{figure}

\subsection{Rule-based Explanations}
An example rule-based explanation on the loan request approval could be ``{\tt Approve} if ${\tt Salary} > 50k$ {\tt AND} ${\tt Age} > 24$'', as shown in \cref{fig:explanations} and \cref{fig:interacting-with-rule-explanations}. Rule-based explanations decompose the models predictions into simple atomic elements either via decision trees, rule lists or rule sets.   While decision trees are similar to rule lists and sets in terms of how they logically represent the decision processes, it is not always the case that decision trees are interpretable from a user perspective, particularly when the number of nodes are large \citep{narayanan2018humans}. As a result, for explanations to facilitate user interaction, the size and complexity of the number of rules and clauses need to be considered.  

LORE~\citep{guidotti2018local} is an example of a local rule-based explainer, which builds an interpretable predictor by first generating a balanced set of neighbour instances of a given data instance through an ad-hoc genetic algorithm, and then building a decision tree classifier out of this set. From this classifier, a local explanation is then extracted, where the local explanation is a pair of logical rules, corresponding to the path in the tree and a set of counterfactual
rules, explaining which conditions need to be changed for the data instance to be assigned the inverted class label by the underlying ML model. Both LIME~\citep{ribeiro2016should}, and Anchors~\citep{ribeiro2018anchors} are other examples which uses a similar intuition such that, even if the decision boundary for the black box ML model can be arbitrarily complex over the whole data space for a given data instance, there is a high chance that the decision boundary is clear and simple in the neighborhood of a data point which can be captured by an interpretable model. Nanfack et al. learn a global decision rules explainer using a co-learning approach to  distill the knowledge of a blackbox model into the decision rules \citep{nanfack21a}.

Rule-based surrogates have the advantage of being interpretable however, in order to achieve coverage, the model must add rules to cover increasingly narrow slices which can in turn negatively impact interpretability. BRCG \citep{dash2018boolean} and GLocalX  \citep{setzu2021glocalx}  trade-off fidelity with interpretability along with compactness of the rules to produce a rule-based representation of the model, which makes them more appropriate as the basis for user feedback.

\subsection{Interacting via Rule-based Explanations}

Rule-based explanations are logical representations of a machine learning model's decision making process which have the benefit of being interpretable to the user. This logical representation can then be modified by end users or domain experts, and these edits bring the advantage that the expected behaviour is more predictable to the end user. Another key advantage of rule-based surrogate models is that, they provide a global representation of the underlying model rather than local explanations which make it more difficult for a user to build up an overall mental model of the system. Additionally, if the local explanations presented to the user to learn and understand the machine learning model are not well distributed, they may create a biased view of the model, leading users to trust a model that for some regions has inaccurate logic. 

\citet{lakkaraju2016interpretable} produce decision sets  where the decision set effectively becomes the predictive model. Popordanoska et al. supports user feedback by generating a rule-based system from data that the user may modify which is then used as a rule-based executable model \citep{popordanoska2020machine}.

\citet{daly2021user} present an algorithm which brings together a rule-based surrogate derived by \citet{dash2018boolean} and an underlying machine learning model to support user provided modifications to existing rules in order to incorporate user feedback in a post-processing approach. Similar to CAIPI~\citep{teso2019explanatory,schramowski2020making} predictions and explanations are presented to experts, however the explanations are rule-based. The user can then specify changes to the labels and rules by adjusting the clauses of the explanation. For example, the user can modify a value such as change ${\tt Age} = 26$ to ${\tt Age} = 30$, remove a clause such as ${\tt Gender} = {\tt male}$ or add a clause such as an additional feature requirement such as ${\tt Employed} = {\tt true}$. The modified labels and clauses are stored together with transformations that map between original and feedback rules as an Overlay or post-processing approach to the existing model. In a counterfactual style approach, the transformation function is applied to relevant new instances presented to the system in order to understand if the user adjustment influences the final prediction. One advantage of the combined approach is that the rule-based surrogate does not need to encode the full complexity of the model as the underlying prediction still comes from the ML model. The rule-based explanations are used as the source of feedback to provide corrections to key variables. Results showed this approach can support updates and edits to an ML model without retraining as a 'band-aid', however once the intended behaviour diverges too much from the underlying model, results deteriorate. \citeauthor{alkan2022frote} similarly use rules as the unit for feedback where the input training data is pre-processed in order to produce an ML model that aligns with the user provided feedback rules \citep{alkan2022frote}. The FROTE algorithm generates synthetic instances that reflect both the feedback rules as well as the existing data and has the advantage of encoding the user feedback into the model. As with the previous solution, the advantage is the rules do not need to reflect the entire model, but can focus on the regions where feedback or correction is needed. The results showed that the solution can support modifications to the ML model even when they diverge quite significantly from that of the existing model. \citet{ratner2017snorkel} combines data-sources to produce weak labels and one source of labels they consider are user provided rules or labelling functions which can then be tuned. 
A recent work~\citep{Lijie2022} explored an explanation-driven interactive machine learning (XIML) system with the Tic-Tac-Toe game as a use case to understand how an XIML mechanism effects users’ satisfaction with the underlying ML solution. A rule based explainer, BRCG \citep{dash2018boolean}, is used for generating explanations, and authors experimented on different modalities to support user feedback through visual or rule-based corrections. They conducted a user study in order to test the effect of allowing users to interact with the rule-based system through creating new rules or editing or removing existing rules. The results of the user study demonstrated that  allowing interactivity within the designed XIML system leads to increased satisfaction for the end users.

\subsection{Benefits and limitations}

An important advantage of leveraging rules to enable user feedback is that, editing a clause can impact many different data points which can aid in reducing the cognitive load for the end user and making the changes more predictable. When modifying a Boolean clause, the feedback and the intended consequences are clearer in comparison to alternative techniques such as re-weighting feature importance or a training data point. However, one challenge with rule based methods is that, feedback can be conflicting in nature, therefore some form of conflict resolution is needed to be implemented in order to allow experts to resolve collisions~\citep{alkan2022frote}. Additionally, rules can be probabilistic and eliciting such probabilities from experts can be difficult. 

An additional challenge is that, most of the existing rule-based solutions only build upon original features in the data. However, recent work has started to consider this direction for example \citep{alaa2019demystifying} produces a symbolic equation as a white-box model and \citep{santurkar2021editing} supports editing concept based rules for image classification. In the next section, let us therefore continue with a branch of work that potentially allows for interactions on a local and global level via concept-based explanations.
\section{Interacting Using Concept-Based Explanations}
\label{sec:concept-based-interaction}

An advantage of input attributions is that they can be extracted from any black-box model without the need for retraining, thus leaving performance untouched.
Critics of these approaches, however, have raised a number of important issues~\citep{rudin2019stop}.
Perhaps the most fundamental ones, particularly from an interaction perspective, are the potential lack of faithfulness of post-hoc explanations and that input-based explanations are insufficient to accurately capture the reasons behind a model's decision, particularly when these are abstract~\citep{stammer2021right,viviano2021saliency}.  Consider an input attribution highlighting a red sports car:  does the model's prediction depend on the fact that a car is present, that it is a sports car, or that it is red?  This lack of precision can severely complicate interaction and revision.

A possible solution to this issue is to make use of white-box models, which -- by design -- admit inspecting their whole reasoning process.  Models in this category include, e.g., shallow decision trees~\citep{angelino17corel,rudin2019stop} and sparse linear models~\citep{ustun2016supersparse} based on human-understandable features.  These models, however, do not generally support representation learning and struggle with sub-symbolic data.

A more recent solution are concept-based models (CBMs), which combine ideas from white and black-box models to achieve partial, selective interpretability. 
Since it is difficult -- and impractical -- to make every step in a decision process fully understandable, CBMs generally break down this task into two levels of processing:
a bottom level, where one module (typically black-box) is used for extracting higher-level concept vectors $c_j(\vx)$, with $j = 1, \ldots, k$, from raw inputs, and
a more transparent, top level module in which a decision $y = f(c_1(\vx), \ldots, c_k(\vx))$ is made \emph{based on the concepts alone}. Most often, the top layer prediction is obtained by performing a weighted aggregation of the extracted concepts. \cref{fig:explanations} provides an example of such concept vectors extracted from the raw data in the context of loan requests. Such concept vectors are often of binary form, e.g. a person applying for a loan is either considered a professional or not. The explanation finally corresponds to importance values on these concept vectors.

CBMs combine two key properties.  First, the prediction is (roughly) independent from the inputs given the concept vectors.  Second, the concepts are chosen to be as human understandable as possible, either by designing them manually or through concept learning, potentially aided by additional concept-level supervision.
Taken together, these properties make it possible to faithfully explain a CBMs predictions based on the concept representation alone, thus facilitating interpretability without giving up on representation learning.
Another useful feature of CBMs is that they allow for \textit{test-time interventions} to introspect and revise a model's decision based on the individual concept activations~\citep{koh2020concept}.

Research on CBMs has explored different representations for the higher-level concepts, including
(i) autoencoder-based concepts obtained in an unsupervised fashion and possibly constrained to be independent and dissimilar from each other~\citep{alvarez2018towards}.
(ii) prototype representations that encode concrete training examples or parts thereof~\citep{chen2019looks, hase2019interpretable, rymarczyk2020protopshare, nauta2021neural, barnett2021iaia},
(iii) concepts that are explicitly aligned to concept-level supervision provided upfront~\citep{koh2020concept, chen2020concept},
(iv) white-box concepts obtained by interactively eliciting feature-level dependencies from domain experts~\citep{lage2020learning}.

The idea and potential of symbolic, concept-based representations is also found in neuro-symbolic models, although this branch of research was developed from a different standpoint than interpretability alone. Specifically, neuro-symbolic models have recently gained increased interest in the AI community~\citep{garcez2012neural, GarcezGLSST19, Yi0G0KT18, wagner2021neural} due to their advantages in terms of performance, flexibility and interpretability. The key idea is that these approaches combine handling of sub-symbolic representations with human-understandable, symbolic latent representations.
Although it is still an open debate on whether neuro-symbolic approaches are ultimately preferable over purely subsymbolic or symbolic approaches, several recent works have focused on the improvements and richness of neuro-symbolic explanations in the context of understandability and the possibilities of interaction that go beyond the approaches of the previous sections. Notably, the distinction between CBMs and neuro-symbolic models can be quite fuzzy, with CBMs possibly considered as one category of neuro-symbolic AI~\cite{SarkerZEH21, YehKR21}.

Although research on concept-level and neuro-symbolic explainability is a flourishing branch of research, it remains quite recent and only selected works have incorporated explanations in an interactive setting.  In this section we wish to provide details on these works, but also mention noteworthy works that show potential for leveraging explanations in human machine interactions.

\subsection{Interacting with Concept-based Models}

\begin{figure}[!t]
    \centering
    \begin{tabular}{c|c}
        \includegraphics[width=0.45\textwidth]{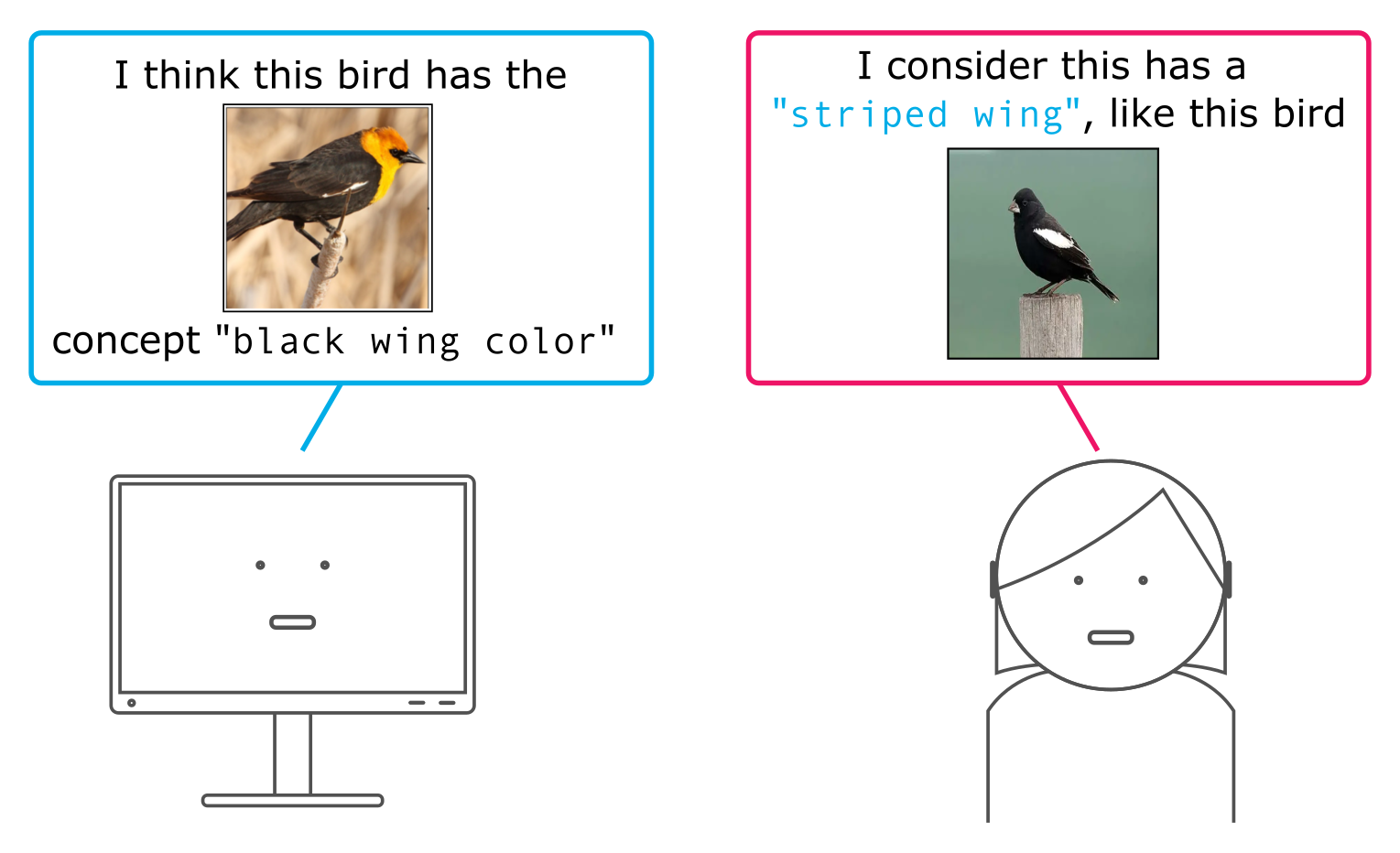}
        & \includegraphics[width=0.45\textwidth]{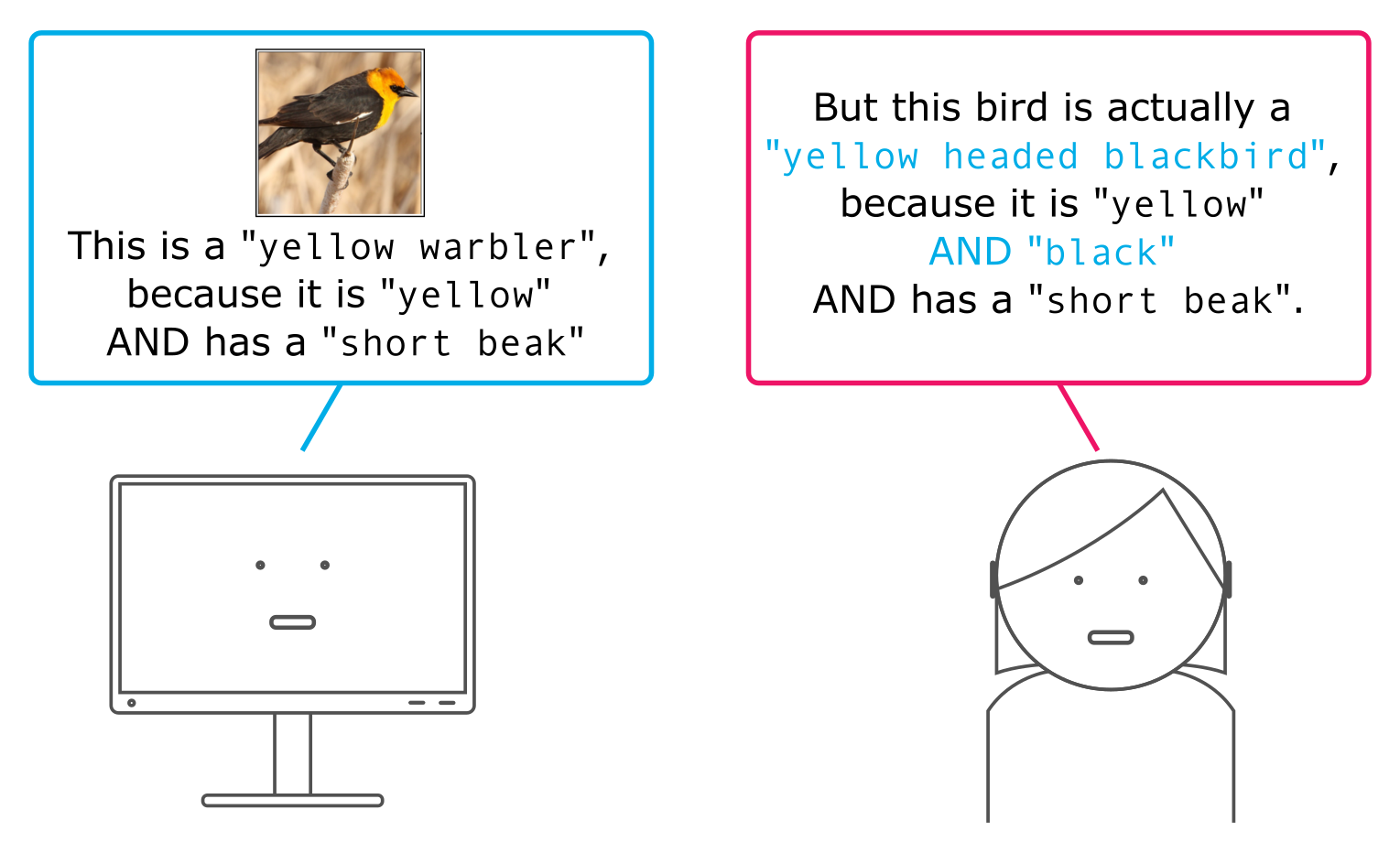}
    \end{tabular}
    \caption{Illustration of providing feedback on concept learning and concept aggregation strategies.
    \textbf{Left}: In concept learning a model learns a set of basic concepts that are present in the dataset. Hereby it learns to grounding specific features of the dataset on symbolic concept labels. 
    \textbf{Right}: Given a set of known basic concepts a model could falsely aggregate the concept activations leading to a false final prediction. Users can provide feedback on the concept explanations.}
    \label{fig:interacting-with-concept-explanations}
\end{figure}

Like all machine learning models, CBMs are also prone to erroneous behaviour and may require revision and debugging by human experts~\citep{bahadori2021debiasing}.
The special structure of CBMs poses challenges that are specific to this setting.
A major issue that arises in the context of CBMs is that not only the weights used in the aggregation of the concept activations can be faulty and require adjustment, but the concepts themselves -- depending on how they are defined or acquired -- can be insufficient, incorrect, or uninterpretable~\citep{lage2020learning,kambhampati21linguafranca}.

These two steps have mostly been tackled separately. \cref{fig:interacting-with-concept-explanations} gives a brief sketch of this where a human user can guide a model in learning the basic set of concepts from raw images (left), but also provide feedback on the concept aggregations, e.g. in case relevant concepts are ignored for the final class prediction (right).

Several works have tackled interactively learning concepts. For instance, \citet{lage2020learning} focus on human-machine concept alignment, and propose an interactive loop in which a human expert directly guides concept learning.  Specifically, the machine elicits potential dependencies between inputs and concepts by asking questions like, e.g., ``does the input variable \texttt{lorazepam} contribute to the concept \texttt{depression}?''.  
This is reminiscent of FIND~\citep{lertvittayakumjorn2020find}, but the queried dependencies are chosen by the machine so to be as informative as possible. In their recent work, \citet{stammer2022interactive} propose to use prototype representations for interactively learning concepts and thereby grounding features of image data to discrete (symbolic) concept representations.  Via these introspectable encodings a human user can guide the concept learning by directly giving feedback on the prototype activations or by providing paired samples that possess specific concepts which the model should learn. The sample pairing feedback is reminiscent of Shao \etal~\citep{shao2022right} who proposed debiasing generative models via weak supervision. \citet{bontempelli2022concept} on the other hand propose debugging part-prototype networks via interactive supervision on the learned part-prototypes using a simple interface in which the user can determine signal that concept is valid or not valid for a particular precision using a single click.  They also remark that this kind of concept-level feedback on concepts generalizes is very rich, in that it naturally generalizes to all instances that feature those concepts, facilitating debugging from few interaction rounds.

Other works focus on the concept aggregation step.
For instance, \citet{teso2019toward} applied explanatory interactive learning to self-explainable neural networks, enabling end-users to fine-tune the aggregation weights~\citep{alvarez2018towards}.
An interesting connection to causal learning is made by \citet{bahadori2021debiasing}, who present an alternative and principled approach for debiasing CBMs from confounders.
These two strategies, however, assume the concepts to be given and fixed, which is often not the case.

Finally, \citet{bontempelli2021toward} outline a unifying framework for debugging CBMs that clearly distinguishes between bugs in the concept set and bugs in the aggregation step, and advocate a multi-step procedure that encompasses determining where the source of the error lies and providing corrective supervision accordingly.

\subsection{Interacting with Neuro-symbolic Models}

Neuro-symbolic models, similar to CBMs, also support communicating with users in terms of symbolic, higher-level concepts as well as more general forms of constraints and knowledge.  There are different ways in which these symbols can be presented and the interaction can be structured.
 
Ciravegna \etal~\citep{ciravegna2020human}, propose to integrate deep learning with expressive human-driven first-order logic (FOL) explanations. Specifically, a neural network maps to a symbolic concept space and is followed by an additional network that learns to create FOL explanations from the latent concept space. Thanks to this direct translation into FOL explanations, it is in principle easy to integrate prior knowledge from expert users as constraints onto the model.

On the other hand Stammer \etal~\citep{stammer2021right} focus more on human-in-the-loop learning. With their work, the authors show that receiving explanations on the level of the raw input -- as done by standard approaches presented in the previous sections -- can be insufficient for removing Clever Hans behavior, and show how this problem can be solved by integrating and interacting with rich, neuro-symbolic explanations. Specifically, the authors incorporate a right for the right reason loss~\citep{ross2017right} on a neural reasoning module which receives multi-object concept activations as input. The approach allows for concept-level explanations, and user feedback is cast in terms of relational logic statements.

The benefits of symbolic explanations and feedback naturally extend from supervised learning and computer vision domains, as in the previous works, to settings like planning~\citep{chakraborti2019plan} and, most recently, reinforcement learning (RL).  In particular, in the context of deep RL, Guan \etal~\citep{guan2021widening} provide coarse symbolic feedback in the form of object-centric image regions to accompany binary feedback on an agent's proposed actions. Another interesting use of symbolic explanations in RL is that of Zha \etal~\citep{zha21ambdemonstrations}, in which an RL agent learns to better understand human demonstrations by grounding these in human-aligned symbolic representations.

\subsection{Benefits and Limitations}

The common underlying motivation of all these approaches is the potential of symbolic communication to improve both precision and bandwidth of explanatory interaction between (partially sub-symbolic) AI models and human stakeholders.
A recent paper by Kambhampati \etal~\citep{kambhampati21linguafranca} provides an excellent motivation on the importance of this property as well as important remaining challenges. The main challenge of concept-based and neuro-symbolic models lies in identifying a set of basic concepts or symbols~\citep{YehKR21} and grounding a model's latent representations on these. Though recent works, e.g. \citet{lage2020learning} and \citet{ stammer2022interactive}, have started to tackle this it remains an important issue to solve particularly for real world data.
\section{Open Problems}
\label{sec:open-problems}

Despite recent progress on integrating explanations and interaction, many unresolved problems remain.
In this section, we outline a selection of urgent open issues and, wherever possible, suggest possible directions forward, with the hope of spurring further research on this important topic.

\subsection{Handling Human Factors}
\label{sec:human-factors}

Machine explanations are only effective insomuch as they are understood by the person at the receiving end.
Simply ensuring algorithmic \textit{transparency}, which is perhaps the major focus in current XAI research, gives few guarantees, because understanding strongly depends on human factors such as mental skills and familiarity with the application domain~\citep{sokol2020one,liao2021human}.
As a matter of fact, factual but ill-designed machine guidance can actively harm understanding~\citep{ai2021beneficial}.

Perhaps the most critical element for successful explanation-based interaction requires is that the user and the machine to agree on the \textit{semantics} of the explanations they exchange.
This is however problematic, partly because conveying this information to users is non-trivial, and partly because said semantics are often unclear or brittle to begin with, as discussed in~\cref{sec:faithfulness-and-debugging,sec:abstraction}.
The literature does provide some guidance on what classes of explanations may be better suited for IML.  Existing studies suggest that people find it easier to handle explanations that express concrete cases~\citep{kim2014bayesian} and that have a counterfactual flavour~\citep{wachter2017counterfactual}, and that breaking larger computations into modules facilitates simulation~\citep{lage2019human}, but more work is needed to implement and evaluate these suggestions in the context of explanation-based IML.
Moreover, settings in which users have also to \textit{manipulate} or \textit{correct} the machine's explanations, like interactive debugging, impose additional requirements.
Another key issue is that of cognitive biases.  For instance, human subjects tend to pay more attention to affirmative aspects of counterfactuals~\citep{byrne2019counterfactuals}, while AIs have no such bias.
Coping with these human factors requires to design appropriate interaction and incorporation strategies, and it is necessary for correct and robust operation of explanation-based IML.

We also remark that different stakeholders inevitably need different kinds of explanations~\citep{miller2019explanation,liao2021human}.
A promising direction of research is to enable users to customize the machine's explanations to their needs~\citep{sokol2020one,finzel2021explanation}.
Challenges on this path include developing strategies for eliciting the necessary feedback and assisting users in exploring the space of alternatives.

\subsection{Semantics and Faithfulness}
\label{sec:faithfulness-and-debugging}

%
%
Not all kinds of machine explanations are equally intelligible and not all XAI algorithms are equally reliable.
For instance, some gradient-based attribution techniques fail to satisfy intuitive properties (like implementation invariance~\citep{sundararajan2017axiomatic}) or ignore information at the top layers of neural networks~\citep{adebayo2018sanity}, while sampling-based alternatives may suffer from high variance~\citep{zhang2019should,teso2019toward}.  A number of other issues have been identified in the literature~\citep{hooker19roar,kindermans2019reliability,adebayo20debugging,sixt2020explanations,kumar2020problems}.
The semantics of transparent models is also not always well-defined.  For instance, the coefficients of linear models are often viewed as capturing feature importance in an additive manner~\citep{ribeiro2016should}, but this interpretation is only valid as long as the input features are independent, which is seldom the case in practice.  Decision tree-based explanations have also received substantial scrutiny~\citep{izza2020explaining}.

%
%
Another critical element is faithfulness.  The reason is that bugs identified by unfaithful explanations may be artifacts in the explanation rather than actual issues with the model, meaning that asking users to correct them is not only deceptive, but also uninformative for the machine and ultimately wasteful~\citep{teso2019explanatory}.
%
The \textit{distribution} of machine explanations is equally important for ensuring faithfulness:  individually faithful local explanations that fail to cover the whole range of machine behaviors may end up conveying an incomplete~\citep{lertvittayakumjorn2020find} and deceptively optimistic~\citep{popordanoska2020machine} picture of the model's logic to stakeholders and annotators.

Still, some degree of unfaithfulness is inevitable, for both computational and cognitive reasons.
On the one hand, interaction should not overwhelm the user~\citep{kulesza2015principles}.  This entails presenting a necessarily simplified picture of the (potentially complex) inference process carried out by the machine.
On the other hand, extracting faithful explanations often comes at a substantial computational cost~\citep{van2021tractability}, which is especially problematic in interactive settings where excessive repeated delays in the interaction can estrange the end-user.  Alas, more light-weight XAI strategies tend to rely on approximations and leverage less well-defined explanation classes.

\subsection{Abstraction and Explanation Requirements}
\label{sec:abstraction}

Many attribution methods are restricted to measuring relevance of individual input variables or training examples.
In stark contrast, explanations used in human--human communication convey information at a more abstract, conceptual level, and as such enjoy improved expressive power.
An important open research question is how to enable machines to communicate using such higher-level concepts, especially in the context of the approaches discussed in~\cref{sec:concept-based-interaction}.

This immediately yields the issue of obtaining a relevant, user-understandable symbolic concept space~\citep{kambhampati21linguafranca,rudin2022challenges}.
This is highly non-trivial.
In many cases it might not be obvious what the relevant higher-level concepts should be, and more generally it is not clear what properties should be enforced on these concepts -- when learned from data -- so to encourage interpretability.  Existing candidates include similarity to concrete examples~\citep{chen2019looks}, usage of generative models~\citep{karras19stylegan}, and enforcing disentanglement among concepts~\citep{scholkopf2021toward,stammer2022interactive}.
One critical challenge is that imperfections in the learned concepts may compromise the predictive power of a model as well as the the semantics of explanations while being hard to spot~\citep{nauta2020looks,hoffmann2021looks,kraft2021sparrow,mahinpei2021promises,margeloiu2021concept}, calling for the development of concept-level debugging strategies~\citep{bontempelli2021toward}.
Additionally, assuming a basic set of concepts has been identified, it seems likely that this set will not be sufficient and should allow for expanding~\citep{kambhampati21linguafranca}.

On the broader topic of explanation requirements, \citet{liao2021human} discuss many different aspects that XAI brings, such as the \textit{diverse explainability needs} of stakeholders due to the no “one-fits-all” solutions from the collection of XAI algorithms, and \textit{pitfalls of XAI} in the sense that there can be a gap between algorithmic explanations that several XAI works provide and the actionable understanding that these solutions can facilitate. One important statement that is highlighted in~\citep{liao2021human} is that, closing the gap between \textit{technicality of XAI} and the \textit{user’s engagement with the explanations} requires considering possible user interactions with XAI, and operationalizing human-centered perspectives in XAI algorithms. 

This latter point also requires developing evaluation methods that better consider the actual user needs in the downstream usage contexts.  Tis further raises the question of whether ``good'' explanations actually exist and how one can quantify these. One interesting direction forward is to consider explanation approaches in which a user can further query an initial model's explanation similar to how humans provide additional (detailed) queries in case the initial explanation is confusing or insufficient.

\subsection{Modulating and Manipulating Trust}
\label{sec:modulating-and-manupulating-trust}

In light of the ethical concern of deploying ML models in more real-world applications, the field of \textit{trustworthy ML} has grown, which studies and pursues desirable qualities such as fairness, explainability, transparency, privacy and robustness~\citep{Varshney2019}. As discussed by \citet{liao2021human}, explainability has moved beyond providing details to comprehend the ML models being developed, and it has rather become an essential requirement for people to trust and adopt AI and ML solutions.

However, the relationship between (high-quality) explanations and trust is not straightforward.  One reason is that explanations are not the only contributing factor~\citep{wang2018my}.
However, while user studies support the idea that explanations enable stakeholders to \textit{reject} trust in misbehaving models, the oft stated claim that explanations help to \textit{establish} trust into deserving models enjoys less empirical support.
This is related to a trend, observed in some user studies, that participants may put too much trust in AI models to begin with~\citep{schramowski2020making}, thus making the effects of additional explanations less obvious.
Understanding also plays a role.  Failure to understand an explanation may drive users to immediately and unjustifiably distrust the system~\citep{honeycutt2020soliciting} and, rather surprisingly, in some cases the mere fact of being exposed to the machine's internal reasoning may induce a loss of trust~\citep{honeycutt2020soliciting}.  More generally, the link between interaction and trust is under-explored.

Another important issue that explanations can be intentionally manipulated by malicious parties so to persuade stakeholders into trusting unfair or buggy systems~\citep{dombrowski2019explanations,heo2019fooling,anders2020fairwashing}.
This is a serious concern for the entire enterprise of explainable AI and therefore for explanatory interaction.
Despite initial efforts on making explanations robust against such attacks, more work is still needed to guarantee safety for models used in practice.

\subsection{Annotation Effort and Quality}
\label{sec:annotation-effort-and-quality}

Explanatory supervision comes with its own set of challenges.
First and foremost, just like other kinds of annotations, corrections elicited from human supervisors are bound to be affected by noise, particularly when interacting with non-experts~\citep{frenay2014classification}.
Since explanations potentially convey much more information than simple labels, the impact of noise is manifold.
One option to facilitate ensuring high-quality supervision is that of providing human annotators with guidance, cf.~\citep{cakmak2014eliciting}, but this cannot entirely prevent annotation noise.
In order to cope with this, it is critical to develop learning algorithms that are robust to noisy explanatory supervision.  An alternative is to leverage interactive strategies for enabling users to identify and rectify mislabeled examples identified by the machine~\citep{zeni2019fixing,teso2021interactive}.

Some forms of explanatory supervision -- for instance, pixel-level relevance annotations -- require higher effort on the annotator's end.
This extra cost is often justified by the larger impact that explanatory supervision has on the model compared to pure label information, but in most practical application effort is constrained by a tight budget and must be kept under control.
Analogously to what is normally done in interactive learning~\citep{settles2012active}, doing so involves developing querying strategies that only eliciting explanatory supervision when strictly necessary (\eg when label information is not enough) and to efficiently identify the most informative queries.  To the best of our knowledge, this problem space is completely unexplored.

\subsection{Benchmarking and Evaluation}

Evaluating and benchmarking current and novel approaches that integrate explanations and interaction is particularly challenging. There are several reasons for this. One reason is the effort in providing extensive user studies, where many recent studies tend to focus on simulated user interactions for evaluations. The difficulties for this are not just the participant organization and proper study design itself (which can lead to many pitfalls if not properly devised), but also from the engineering perspective a swift user feedback integration is not immediate in all studies. Thus, in many cases additional engineering and an extensive user study remain necessary before real-world deployment. 

Further reasons are the individual use case of a method, but also the possibly high variance in user's feedback which make it challenging to assess a methods properties with one task and study alone. A very important branch for future research is thus to develop more standardized benchmarking tasks and metrics for evaluating
such methods, where \citet{friedrich2022typology} provide an initial set of important evaluation metrics and tasks for future research on XIL.
\section{Related Topics}
\label{sec:related-topics}

%
%
Research on integrating interaction and explanations builds on and is related to a number of different topics.
The main source of inspiration is the vast body of knowledge on explainable AI, which has been summarized in several overviews and surveys~\citep{guidotti2018survey,ras2022explainable,liao2021human,gilpin2018explaining,montavon2018methods,adadi2018peeking,carvalho2019machine,sokol20factsheets,belle2021principles}.
A major difference to this literature is that in XAI the communication between machine and user stops after receiving the machine's explanations.

Recommender systems are another area where explanations have found wide applicability, with a large number of approaches being proposed and applied in real-world systems in recent years.  Compatibly with our discussion, explanations has been shown to improve transparency~\cite{Rashmi2002, Nina2015} and trustworthiness~\cite{Zhang2018} of recommendations, to help users make better decisions, and to enable users to provide feedback to the system by correcting any incorrect assumptions that the recommender has made for their interests~\cite{Alkan2019}. With critiquing-based recommenders, users can critique the presented suggestions and provide their preferences~\cite{Antognini2021InteractingWE, Wu2019critique, chen2012critiquing}.  Here we focus on parallel advancements made in interactive ML, and refer the reader to \citet{zhang2020explainable} for a comprehensive review of explainable recommender systems.

The idea of using explanations as supervision in ML can be traced back to explanation-based learning~\citep{mitchell1986explanation,dejong1986explanation}, where the goal is to extract logical concepts by generalizing from symbolic explanations, although the overall framework is not restricted to purely logical data.

Another major source of inspiration are approaches for \textit{offline} explanation-based debugging, a topic has recently received substantial attention, especially in natural language processing community~\citep{lertvittayakumjorn2021explanation}.  This topic encompasses, for instance, learning from annotator rationales~\citep{zaidan2007using}, i.e., from snippets of text appearning in a target sentence that support a particular decision, and that effectively the same role as input attributions.  Recent works have extended this setup to complex natural language inference tasks~\citep{camburu2018snli}.
Another closely related topic is learning from feature-level supervision, which has been used to complement standard label supervision for learning higher quality predictors~\citep{decoste2002training,raghavan2006active,raghavan2007interactive,druck2008learning,druck2009active,attenberg2010unified,small2011constrained,settles2011closing}.  These earlier approaches assume (typically complete) supervision about attribute-level relevance to be provided in advance, independently from the model's explanations, and focus on shallow models.
More generally, human explanations can be interpreted as encoding prior knowledge and thus used guide the learning process towards better aligned models faster~\citep{hase2021can}.  This perspective overlaps with and represents a special case of informed ML~\citep{vonruede21informed,beckh2021explainable}.

Two other areas where explanations have been used to inform the learning process are explanation-based distillation and regularization.
The aim of distillation is to compress a larger model into a smaller one for the purpose of improving efficiency and storage costs.  It was shown that paying attention to the explanations of the source model can tremendously (and provably) improve the sample complexity of distillation~\citep{milli2019model}, hinting at the potential of explanations as a rich source of supervision.
Regularization instead aims at encouraging models to produce more simulatable~\citep{wu2018beyond,wu2020regional} or faithful explanations~\citep{plumb2020regularizing} by introducing an additional penalty into the learning process.
Here, however, no explanatory supervision is involved.

%
%
Two types of explanations that we have not delved into are counterfactuals and attention.
Counterfactuals identify changes necessary for achieving alternative and possibly more desirable outcomes~\citep{wachter2017counterfactual}, for instance what should be changed in a loan application in order for it to be approved.  They have become popular in XAI as a mean to help stakeholders to form actionable plans and control the system's behavior~\citep{lim2012improving,karimi2021algorithmic}, but have recently shown promise as a mean to design novel interaction strategies~\citep{kaushik2019learning,wu2021polyjuice,de2022generating}.
Attention mechanisms~\citep{bahdanau2014neural,vaswani2017attention} also offer insight into the decision process of neural networks and, although their interpretation is somewhat controversial~\citep{bastings2020elephant}, they are a viable alternative to gradient-based attributions for integrating explanatory feedback into the model in an end-to-end manner~\citep{mitsuhara2019embedding,heo2020cost}.


%
%
Finally, another important aspect that we had to omit is causality~\citep{pearl2009causality}, which is perhaps the most solid foundation for imbuing cause-effect relationships within ML models~\citep{chattopadhyay2019neural,xu2021causality}, and conversely the ability to identify causal relationships between inputs and predictions constitutes a fundamental step towards explaining model predictions~\citep{Holzinger2019,geiger2021causal}.
Work on causality in interactive ML is however sparse at best.
\section{Conclusion}

This overview provides a conceptual guide on current research on integrating explanations into interactive machine learning for the purpose of establishing a rich bi-directional communication loop between machine learning models and human stakeholders in a way that is beneficial to all parties involved.
Explanations make it possible for users to better understand the machine's behavior, spot possible limitations and bugs in its reasoning patterns, establish control over the machine, and modulate trust.  At the same time, the machine obtains high-quality, informed feedback in exchange.
We categorized existing approaches along four dimensions, namely algorithmic goal, type of machine explanations involved, human feedback received, and incorporation strategy, facilitating the identification of links between different approaches as well as respective strengths and limitations.
In addition, we identified a number of open problems impacting the human and machine sides of explanatory interaction and highlighted noteworthy paths of future, with the goal of spurring further research into this novel and promising approach, helping to bridge the gap towards human-centric machine learning and AI.

\subsection*{Acknowledgements}

The research of ST was partially supported by TAILOR, a project funded by EU Horizon 2020 research and innovation programme under GA No 952215.

\bibliographystyle{abbrvnat}
\bibliography{paper,explanatory-supervision}
\end{document}